\documentclass[11pt,a4paper]{article}
\usepackage[hyperref]{acl2021}
\usepackage{times}
\usepackage{latexsym}

\usepackage{microtype}

\aclfinalcopy 

\title{Text-to-SQL in the Wild: A Naturally-Occurring Dataset Based on Stack Exchange Data}

\author{\makecell{Moshe Hazoom$^{1}$ ~~~~~~~ Vibhor Malik$^{2}$   ~~~~~ Ben Bogin$^{1,3}$ } \\ 
$^{1}$Rupert \hspace{5mm} $^{2}$Columbia University \hspace{5mm} $^{3}$Tel Aviv University\\
\texttt{\makecell{\{ben,moshe\}@hirupert.com, malik.vibhor@columbia.edu \\
}}}

\date{}
  
\newif\ifcomments
\commentstrue
\ifcomments
    \providecommand{\bb}[1]{{\protect\color{olive}{[BB: #1]}}}
    \providecommand{\mh}[1]{{\protect\color{purple}{[MH: #1]}}}
\else
    \providecommand{\bb}[1]{}
    \providecommand{\mh}[1]{}
\fi

\usepackage{graphicx}

\usepackage{makecell}
\usepackage{booktabs}
\usepackage{tabularx}
\usepackage{arydshln}

\usepackage[T1]{fontenc}

\newcommand{\PreserveBackslash}[1]{\let\temp=\\#1\let\\=\temp}
\newcolumntype{C}[1]{>{\PreserveBackslash\centering}p{#1}}
\newcolumntype{R}[1]{>{\PreserveBackslash\raggedleft}p{#1}}
\newcolumntype{L}[1]{>{\PreserveBackslash\raggedright}p{#1}}

\newcommand\sede{\textsc{SEDE}}

\begin{document}
\maketitle
\begin{abstract}
Most available semantic parsing datasets, comprising of pairs of natural utterances and logical forms, were collected solely for the purpose of training and evaluation of natural language understanding systems. As a result, they do not contain any of the richness and variety of natural-occurring utterances, where humans ask about data they need or are curious about. In this work, we release \sede{}, a dataset with 12,023 pairs of utterances and SQL queries collected from real usage on the Stack Exchange website. We show that these pairs contain a variety of real-world challenges which were rarely reflected so far in any other semantic parsing dataset, 
propose an evaluation metric based on comparison of partial query clauses that is more suitable for real-world queries, and conduct experiments with strong baselines, showing a large gap between the performance on \sede{} compared to other common datasets. 
\end{abstract}

\section{Introduction}
Semantic parsing, the task of mapping natural language into logical forms that can be executed on a database or knowledge graph, has been studied mostly on \emph{academic datasets}, where both the utterances and the queries were written as part of a dataset collection process \cite{hemphill-etal-1990-atis,geoquery,yu-etal-2018-spider}, and not in a natural process where users ask questions about data they need or are curious about. As a result, these datasets generally do not contain any of the richness and diversity of natural-occurring utterances, even if the data on which the questions are asked about is collected from a real-world source.

\begin{table}[]
\centering
\small
\begin{tabular}{L{7.5cm}}
\toprule
\textbf{Title}: \emph{Questions which attract bad answers} \\
\textbf{Description}: \emph{posts which have attracted significantly more controversial or bad answers than good ones} \\
\midrule
\scriptsize
\texttt{\makecell[l]{SELECT  p.Id as [Post Link], p.Score from (\\
~SELECT p.ParentId, count(*) as ContACnt from (\\
~~SELECT PostId, \\
~~~up = sum(case when VoteTypeId = 2 then 1 \\
~~~~~else 0 end), \\
~~~down = sum(case when VoteTypeId = 3 then 1 \\
~~~~~else 0 end) \\
~~~FROM Votes v join Posts p on p.Id = v.PostId \\
~~~WHERE VoteTypeId in (2,3) and PostTypeId = 2 \\
~~~group by PostId \\
~) as ContA \\
~JOIN posts p on ContA.PostId = p.Id \\
~WHERE down > (up / \#\#UVDVRatio:int\#\#) and \\
~(down + up) > \#\#MinVotes:int\#\# \\
~GROUP BY p.ParentId \\ 
) as ContQ \\
JOIN posts p on ContQ.ParentId = p.Id \\ 
WHERE ContQ.ContACnt > (p.AnswerCount / 2) and\\
p.AnswerCount > 1 \\
ORDER BY Score desc}} \\
\bottomrule
\end{tabular}
\caption{Example from \sede{} for a title and description given by the user, together with the SQL query that the user has written.}
\label{tab:SEDE_example}
\end{table}

Recent methods \cite{wang-etal-2020-rat,herzig-etal-2020-tapas,yu2021grappa} have significantly improved results on such academic datasets: state-of-the-art models have yield impressive results of over 70\%, for example, on Spider \cite{yu-etal-2018-spider} in a challenging cross-domain setup, where models are trained and tested on different domains, and up to 80\%-90\% \cite{nguyen2021phrasetransformer,zhao2014typedriven} on single-domain datasets such as \textsc{ATIS} \cite{hemphill-etal-1990-atis} and GeoQuery \cite{geoquery}. While the cross-domain, zero-shot setup introduces many generalization challenges such as non-explicit mentioning of column names and domain-specific phrases \cite{suhr-etal-2020-exploring,deng2020structure-grounded}, we argue that even in the easier single-domain setup, it is still unclear how well state-of-the-art models generalize to the challenges that arise from real-world utterances and queries.

In this work, we take a significant step towards evaluation of Text-to-SQL models in a real-world setting, by releasing \sede{}: a dataset comprised of 12,023 complex and diverse SQL queries and their natural language titles and descriptions, written by real users of the Stack Exchange Data Explorer out of a natural interaction.

In Table~\ref{tab:SEDE_example} we show an example for a SQL query from \sede{}, with its title and description. It introduces several challenges that have not been commonly addressed in currently available datasets: comparison between different subsets, complex usage of 2 nested sub-queries and an under-specified question, which doesn't state what \emph{``significantly more''} means (solved in this case with an input parameter, \texttt{\#\#UVDVRation\#\#}).

Compared to other Text-to-SQL datasets, we show that \sede{} contains at least 10 times more SQL queries templates (queries after canonization and anonymization of values) than other datasets, and has the most diverse set of utterances and SQL queries (in terms of 3-grams) out of all single-domain datasets. We manually analyze a sample of examples from the dataset and list the introduced challenges, such as under-specification, usage of parameters in queries, dates manipulation and more.

We also address the challenging problem of evaluating naturally-occurring Text-to-SQL datasets. In academic datasets, standard evaluation metrics such as denotation accuracy and exact comparison of SQL components can often be used with relative success, but we found this to be a greater challenge in \sede{}. Denotation accuracy is inaccurate for under-specified utterances, where any single clause not mentioned in the question could entirely change execution results, while exact match comparison of SQL components (e.g. comparing all \texttt{SELECT}, \texttt{WHERE}, \texttt{GROUP BY} and \texttt{ORDER BY} clauses) are often too strict when queries are highly complex. While solving these issues still remains an open problem, to at least partially address them we propose to measure a softer version of the exact match metric, PCM-F1, based on partially extracted queries components, and show that this metric gives a better indication of models' performance than common metrics, which yield a score that is close to 0.

Finally, we test strong baselines on our dataset, and show that even models that get strong results on Spider's development set (63.2\% Exact-Match, 86.3\% PCM-F1), perform poorly on our dataset, with a PCM-F1 value of 50.6\%. We hope that the unique and challenging properties exhibited in \sede{}\footnote{Our dataset and code to run all experiments and metrics is available at \url{https://github.com/hirupert/sede}.} will pave a path for future work on generalization of Text-to-SQL models in real world setups.
\section{Background}
In the past decades, a broad selection of datasets have been used as benchmarks for semantic parsing: ATIS \cite{hemphill-etal-1990-atis}, GeoQuery \cite{geoquery}, Restaurants \cite{tang2000}, Scholar \cite{iyer2017}, Academic \cite{li2014}, Yelp and IMDB \cite{yaghmazadeh2017}, Advising \cite{finegan-dollak-etal-2018-improving}, WikiSQL \cite{zhong2017seq2sql}, Spider \cite{yu-etal-2018-spider}, WikiTableQuestions \cite{pasupat-liang-2015-compositional}, Overnight \cite{wang-etal-2015-building} and more. However, the utterances and queries in all of these academic datasets, to the best of our knowledge, were collected explicitly for the purpose of evaluating semantic parsing models, usually with the help of crowd-sourcing (even though in most cases questions are asked about real data). As such, these academic datasets were generated in an artificial process, which often introduces various simplifications and artifacts which are not seen in real-life.

\paragraph{Utterance-Query alignment} One arising issue with this artificial process is that utterances are often aligned to their SQL queries counterparts, such that the columns and the required computations are explicitly mentioned \cite{suhr-etal-2020-exploring,deng2020structure-grounded}. In contrast, natural utterances often do not explicitly mention these, since the schema of the database is not necessarily known to the asking user (for example, the question from Spider "titles of films that include 'Deleted Scenes' in their special feature section" might have been more naturally phrased as "films with deleted scenes" in a real-world setting).

\paragraph{Well-specified utterances} Furthermore, the utterances in academic datasets are mostly well-specified, whereas in contrast, natural utterances are often under-specified or ambiguous; they could be interpreted in different ways and in turn be mapped to different SQL queries. Consider the example in Table~\ref{tab:SEDE_example}: the definition of \emph{``bad answers''} is not well-defined, and in fact could be subjective. Since under-specified utterances, by definition, can not always be answered correctly, any human or machine attempting to answer such a question would have to either make an assumption on the requirement (usually based on previously seen examples) or ask follow-up questions in an interactive setting  \cite{yao-etal-2019-model,elgohary2020speak,elgohary-etal-2021-nl}.

\paragraph{Scope} Last, in academic datasets the utterances are usually written by crowd-sourced workers, asked to provide utterances on various data domains which they do not necessarily need or are interested with. As a result, the utterances and queries are often not very diverse or realistic, are inherently limited in scope, and might not reflect real-world utterances.

\begin{table*}[tb]
\vskip -2.4mm
  \small
  \scriptsize
  \centering
  \setlength{\tabcolsep}{3pt}
      \begin{tabular}{m{3.0cm}cccm{3.35cm}lc}
    \toprule
    
    \multicolumn{1}{c}{\bf Category} &
    \multicolumn{3}{c}{\bf Dataset}  &
    \multicolumn{2}{c}{\bf Example test cases} \\
    \cmidrule(lr){2-4}
    \cmidrule(lr){5-6}
    &
    \multicolumn{1}{c}{\textsc{SEDE}} &
    \multicolumn{1}{c}{Spider} &
    \multicolumn{1}{c}{\textsc{ATIS}} &
    \multicolumn{1}{c}{Title} &
    \multicolumn{1}{c}{SQL query}
    \\
    \midrule
    \textbf{Under specification and Hidden assumptions} & 87 & 14 & 15 & \emph{User List: Highest downvotes per day ratio with minimum downvotes} & \texttt{WHERE id <> -1} \\
    \addlinespace
    \textbf{Parameters} & 40 & 0 & 0 & \emph{Rollbacks by a certain user} & \texttt{WHERE UserId = @UserId} \\
    \addlinespace
    \textbf{Window functions} & 8 & 0 & 0 & \emph{List of users in the Philippines.} & \texttt{DENSE\_RANK() OVER (ORDER BY Reputation DESC)} \\
    \addlinespace
    \textbf{Dates manipulation} & 15 & 0 & 0 & \emph{Quickest new contributor answers to new contributor questions} & \texttt{DATEDIFF(s, Q.CreationDate, A.CreationDate)} \\
    \addlinespace
    \textbf{Numerical computations and text manipulation} & 35 & 0 & 0 & \emph{Average Number of Views per Tag} & \texttt{sum(p.ViewCount)/count(*)} \\
    \addlinespace
    \textbf{DECLARE/WITH} & 11 & 0 & 0 & \emph{Rollbacks by a certain user} & \texttt{DECLARE @UserId AS int = \#\#UserId:int\#\#} \\
    \addlinespace
    \textbf{CASE} & 10 & 0 & 0 & \emph{Questions and answers per year} & \texttt{CASE WHEN Score < 0 THEN 1 ELSE 0 END}\\
    \bottomrule
  \end{tabular}
  \vspace{-1mm}
  \caption{Dataset characteristics comparison of randomly selected 100 samples among \textsc{SEDE} and other popular Text-to-SQL datasets.
  \vspace{-2mm}
  } \label{tab:eg:dataset-characteristics-comparison}
\end{table*}

\section{Stack Exchange Data Explorer}
\label{sec:dataset}
To introduce a realistic Text-to-SQL benchmark, we gather SQL queries together with their titles and descriptions from a naturally occurring dataset: the Stack Exchange Data Explorer. Stack Exchange is an online question \& answers community, with over 3 million questions asked. The Data Explorer\footnote{Publicly available at \url{https://data.stackexchange.com/}} allows any user to query the database of Stack Exchange with T-SQL (a SQL variant) to answer any question they are curious about. The database schema\footnote{\url{https://tinyurl.com/sedeschema}} is spread across 29 tables and 211 columns.
Common utterance topics are published posts, comments, votes, tags, awards, etc.

Any query that users run in the data explorer is logged, and users are able to save the queries with a title and description for future use by the public. All of these logs are available online, and Stack Exchange have agreed to release these queries, together with their title, description and other meta-data. We publish our clean version of this log, which contains 12,023 samples, of which a subset of 1,714 examples is verified by humans to be correct and is used for validation and test. In this section, we explain the cleaning process, analyze the characteristics of the dataset and compare it to other semantic parsing datasets.

\subsection{Data cleaning}
The raw aggregated log contains over 1.6 million queries, however in its raw form many of the rows are duplicated or contain unusable queries or titles. The reason for this large difference between the original data size and the cleaned version is that any time that the author of the query executes it, an entry is saved to the log. This introduces two issues: First, many of the queries are not complete, since they were executed before writing the entire query (these incomplete queries are usually valid and executable, but are missing some expressions with respect to the given title and description). Second, after completing the writing of a correct query, users often keep changing and executing the query, but they do not update the title and description accordingly.

To alleviate these issues, we write rule-based filters that remove bad queries/descriptions pairs with high precision. For example, we filter out examples with numbers in the description, if these numbers do not appear in the query (refer to the pre-processing script in the repository for the complete list of filters and the number of examples each of them filter). Whenever a query has multiple versions due to multiple executions, we take the last executed query which passed all filters. After this filtering step, we are left with 12,309 examples.

Using these filters cleans most of the noise, but not all of it. To complete the cleaning process, we manually go over the examples in the validation and test sets, and either filter-out wrong examples or perform minimal changes to either the utterances or the queries (for example, fix a wrong textual value) to ensure that models are evaluated with correct data.
Out of the 2,000 examples that we have evaluated, we have kept 1,024 and fixed 690\footnote{We publish both the original and the fixed examples}, leading to a total of 1,714 validated examples which we use for validation and test. While we do not perform verification on the training set, the verification procedure on the validation set allows us to estimate that most of the queries (85.7\%) are either entirely accurate or need just a minimal change to be entirely accurate. For example, when the utterance is ''users in Brazil'' while the matching query contains the expression: \texttt{WHERE users.location like \%russia\%} we either change the utterance to ''users in russia'' or change the expression to \texttt{WHERE users.location like \%Brazil\%}. The final number of all training, validation and test examples is 12,023.

\subsection{Dataset Characteristics}
\label{subsec:character}
In this sub-section, we quantify and analyze the introduced challenges in \sede{}, compared to other commonly used semantic parsing datasets.

First, we manually analyze a sample of 100 examples from \sede{} and define 7 categories of introduced challenges. To quantify how often each of these concepts appear in \sede{} in comparison to other datasets (\textsc{Spider} and \textsc{ATIS}), we sample a subset of equal size from each of the other datasets and count the appearances of these concepts. The analysis is shown in Table~\ref{tab:eg:dataset-characteristics-comparison}. Next, we describe each of these concepts.

\paragraph{Under specification and Hidden assumptions} Utterances in \sede{} are often under-specified, that is, they could be interpreted in different ways. For example, when users write \emph{``top users''}, they might refer to users with the most reputation, but also to users that have written the most answers. Likewise, when users write \emph{``last 500 posts''} they might expect to get just the title field of the posts, but possibly also IDs and dates. Similarly, query authors often add various \emph{assumptions} to the queries which are not mentioned in the questions, because they require some knowledge of the available data. For example, they might filter out a special ``Community'' user in StackExchange, which should not be accounted for in computation of votes. We consider an utterance/query pair to be under-specified or contain an hidden assumption whenever the query contains an expression in any of the SQL clauses (\texttt{SELECT}, \texttt{WHERE}, etc.) which is \emph{not} specified in the utterance, or where it is specified in an ambiguous way.

\paragraph{Parameters} In some cases, query authors can address under-specified utterances by letting the user fill in the under-specified parameters, which are marked in \sede{} with either two hashtags (\#) on each side of the parameter name, optionally including the required value type (int, string, etc.) and a default value (e.g. \texttt{\#\#UserId:int\#\#}), or using a declared variable using SQL syntax (e.g. \texttt{@UserId}). For example, in Table~\ref{tab:SEDE_example}, the parameter \texttt{\#\#UVDVRatio:int\#\#} is used to indicate that the user should fill in an integer to specify the ratio that \emph{``significantly more''} refers to. More broadly, parameters are also helpful for re-usability, allowing users unfamiliar with a query to effortlessly change some values in it.

\begin{table*}[]
\centering
\scriptsize
\begin{tabular}{lcccccccc}
\toprule
\multicolumn{1}{l}{}
&\multicolumn{1}{c}{Unique}
&\multicolumn{1}{c}{Unique}
&\multicolumn{1}{c}{Average unique}
&\multicolumn{1}{c}{Utterance}
&\multicolumn{1}{c}{SQL}
&\multicolumn{1}{c}{Avg. nesting}
&\multicolumn{1}{c}{Unique}
&\multicolumn{1}{c}{Average unique}
\\
\multicolumn{1}{l}{Dataset}
&\multicolumn{1}{c}{Utterances}
&\multicolumn{1}{c}{Queries}
&\multicolumn{1}{c}{Tables / uttr}
&\multicolumn{1}{c}{3-gram}
&\multicolumn{1}{c}{3-gram}
&\multicolumn{1}{c}{Level}
&\multicolumn{1}{c}{Templates}
&\multicolumn{1}{c}{Queries / template}
\\
\midrule
Spider                  & 8,034 & 4491 & 1.71 & 41.7K & 25.2K & 1.15 & 1,059 & 7.6 \\
WikiSQL                 & \bf{80,654}\textsuperscript{$\dagger$} & \bf{77,840}\textsuperscript{$\dagger$} & 1\textsuperscript{$\dagger$} & \bf{375K} & \bf{209K} & 1\textsuperscript{$\dagger$} & 488\textsuperscript{$\dagger$} & 165.3\textsuperscript{$\dagger$} \\
\hdashline
Academic                  & 196\textsuperscript{$\dagger$} & 185\textsuperscript{$\dagger$} & 3.0\textsuperscript{$\dagger$} & <1K & <1K & 1.04\textsuperscript{$\dagger$} & 92\textsuperscript{$\dagger$} & 2.1\textsuperscript{$\dagger$} \\
Advising                  & 4,570\textsuperscript{$\dagger$} & 211\textsuperscript{$\dagger$} & 3.0\textsuperscript{$\dagger$} & 20K & 11.2K & 1.18\textsuperscript{$\dagger$} & 174\textsuperscript{$\dagger$} & 20.3\textsuperscript{$\dagger$} \\
ATIS                  & 5,280\textsuperscript{$\dagger$} & 947\textsuperscript{$\dagger$} & \bf{3.8}\textsuperscript{$\dagger$} & 13.2K & 5.8K & 1.39\textsuperscript{$\dagger$} & 751\textsuperscript{$\dagger$} & 7.0\textsuperscript{$\dagger$}\\
GeoQuery                 & 877\textsuperscript{$\dagger$} & 246\textsuperscript{$\dagger$} & 1.1\textsuperscript{$\dagger$} & 1.5K & 1.4K & \bf{2.03}\textsuperscript{$\dagger$} & 98\textsuperscript{$\dagger$} & 8.9\textsuperscript{$\dagger$} \\
IMDB                 & 131\textsuperscript{$\dagger$} & 89\textsuperscript{$\dagger$} & 1.9\textsuperscript{$\dagger$} & <1K & <1K & 1.01\textsuperscript{$\dagger$} & 52\textsuperscript{$\dagger$} & 2.5\textsuperscript{$\dagger$} \\
Restaurants                 & 378\textsuperscript{$\dagger$} & 23\textsuperscript{$\dagger$} & 2.3\textsuperscript{$\dagger$} & <1K & <1K & 1.17\textsuperscript{$\dagger$} & 17\textsuperscript{$\dagger$} & 22.2\textsuperscript{$\dagger$} \\
Scholar                 & 817\textsuperscript{$\dagger$} & 193\textsuperscript{$\dagger$} & 3.2\textsuperscript{$\dagger$} & 2.6K & 2.2K & 1.02\textsuperscript{$\dagger$} & 146\textsuperscript{$\dagger$} & 5.6\textsuperscript{$\dagger$} \\
Yelp                 & 128\textsuperscript{$\dagger$} & 110\textsuperscript{$\dagger$} & 1.0\textsuperscript{$\dagger$} & <1K & <1K & 1.0\textsuperscript{$\dagger$} & 89\textsuperscript{$\dagger$} & 1.4\textsuperscript{$\dagger$} \\
\midrule
SEDE                  & 12,023 & 11,767 & 2.14 & 42.6K & 173K & 1.28 & \bf{10,664} & \bf{1.1} \\
\bottomrule
\end{tabular}
\caption{Comparison of different semantic parsing datasets (for Spider, analysis is performed on training and validation sets only). $\dagger$ denotes that numbers are reported from \citet{finegan-dollak-etal-2018-improving}. Average Unique Queries / template denotes the number of different SQL queries per template, thus lower means more diversity in the dataset. Datasets above dashed line are cross-domain, and below it are single-domain.}
\label{tab:datasets}
\end{table*}

\paragraph{Window functions} Window functions operate on a set of rows and return a single value for each row from the underlying query, thus allowing to perform various aggregation operators without the need for a separate aggregation query. Window functions are often used in \sede{} to report percentiles of a specific value in a row, by using operators such as \texttt{ROW\_NUMBER() OVER}, \texttt{NTILE}, \texttt{TOP(X) PERCENT}, etc.

\paragraph{Dates manipulation} Queries in \sede{} sometimes contain dates arithmetic expressions. See the example category query in Table~\ref{tab:eg:dataset-characteristics-comparison}: this expression calculates the difference in seconds from the time the question was created to the time the answer was created.

\paragraph{Numerical computations and text manipulation} Queries can perform any arbitrary numerical computation and text manipulation. The computations in \sede{} often include multiple nested operators including rounding and conversions to float, for example: \texttt{ROUND(CAST(Main.Total AS FLOAT) / Meta.Total, 2) AS 'Ratio'}. Queries can also contain text manipulation such as concatenation, for example: \texttt{'stackoverflow.com/tags/' + t.tagName + '/info' as [Link]} which builds a URL from a tag name.

\paragraph{DECLARE/WITH} 
SQL queries can be written as a procedural process, where multiple commands are executed sequentially. Query authors can store values in simple variables with \texttt{DECLARE}, but more importantly, they can store complete ``views'' of tables with the \texttt{WITH} command. While these commands do not add any expressivity (that is, any query can be written without these commands), they allow writing more clear and concise queries with less nested expressions.

\paragraph{CASE} The \texttt{CASE} clause is similar to an \emph{if-then-else} statement of any programming language, and is often used to either make the query more readable (e.g. by returning names of values instead of integers) or to perform conditional logic. For example, the clause in Table~\ref{tab:eg:dataset-characteristics-comparison} (last row) counts negative scores using \texttt{CASE} function.

\paragraph{Comparison}
In Table~\ref{tab:eg:dataset-characteristics-comparison} we see that a vast majority of \sede{} is not well-specified, which implies that in order for Text-to-SQL models to work robustly in a real-world setting, it should identify cases of ambiguity and possibly proceed with follow-up questions. We see that the rest of the concepts appear in 10\% to 40\% of \sede{} examples, whereas these concepts are not exhibited in any other analyzed dataset.

Next, we show a comparison of quantifiable metrics of popular Text-to-SQL datasets compared to \sede{} in Table~\ref{tab:datasets}. We see that \sede{} is the largest dataset in terms of unique utterances and queries out of all single-domain datasets. To compare diversity and scope, we also measure the number of unique 3-grams for both the utterances and the queries, and see that \sede{} has a very diverse set of SQL 3-grams, with almost 6 times the number of the next follower, Spider, and only 17\% less than WikiSQL, which is 6.6 bigger in terms of queries. The number of utterance 3-gram is the second largest, after WikiSQL. Last, we count the number of unique \emph{SQL templates}, as defined in \citet{finegan-dollak-etal-2018-improving}: we anonymize the values and group all canonized queries. We see that \sede{} has more than 10 times templates than the follower Spider, and that the average number of queries per template is the lowest. We also see that \sede{} is third in terms of average nesting level, after ATIS and GeoQuery.

\subsection{Limitations}
We note that in order to simulate the most realistic setting, an ideal Text-to-SQL dataset would include questions asked by users which are completely unaware of the schema, which are not SQL-savy, and that the person asking the question would be different than the person answering it. While this is not the case in \sede{}, we believe its setting is still significantly more realistic that other datasets.
\section{Evaluation}
Semantic parsing models are usually evaluated in two different forms: execution accuracy and logical forms accuracy. In this section, we show why using any of these metrics is difficult with complex queries such as those in \sede{}, and propose a more loose metric for evaluation of models.

\paragraph{Execution accuracy} This metric is measured by executing both the predicted and gold query against a dataset, and considers the query to be correct if the two output results are the same (or similar enough). While this metric appears to be exactly what we want to optimize (yielding a query the outputs a correct output), it does not necessarily cope well with two challenges: spurious queries and under-specified questions. Spurious queries are incorrect queries (with respect to the given question) that happen to result in a correct answer, thus leading to a false-positive count. The problem of spuriousness can be addressed by executing the predicted query on modified versions of the dataset, as proposed in \citet{zhong-etal-2020-semantic-evaluation}. The second challenge, evaluating under-specification, is arguably harder to address, as mentioned in Subsection~\ref{subsec:character}. For example, consider a question that asks for \emph{"the top 1\% active users"}. This question does not specify which columns should be returned, how the rows should be ordered, and how does one measure ``being active''. As such, a query could be correct with respect to some interpretation, yet its execution result might be different than the execution result of the given gold query.

\paragraph{Logical form accuracy} Instead of comparing execution results, another frequent approach is to simply perform a textual comparison between the predicted and gold queries. When comparing SQL queries, it is common to perform a more loose comparison that does not consider the order of appearances of different clauses (e.g. it shouldn't matter which \texttt{WHERE} expression is written first), as performed in Spider \cite{yu-etal-2018-spider}. However, as discussed in \citet{zhong-etal-2020-semantic-evaluation}, even this looser metric leads to false-negative measures, since multiple queries can all be correct with respect to an utterance, but written in various different manners. Due to the richness of SQL queries in \sede{}, its extended scope and the fact that queries are written by many different authors, in our case this problem deteriorates: queries can be written in a substantial number of ways. For example, a query that contains a \texttt{WITH} statement could yield exactly the same result without it, by including a nested \texttt{FROM} clause instead.

\begin{figure}
  \centering
  \includegraphics[width=0.7\linewidth]{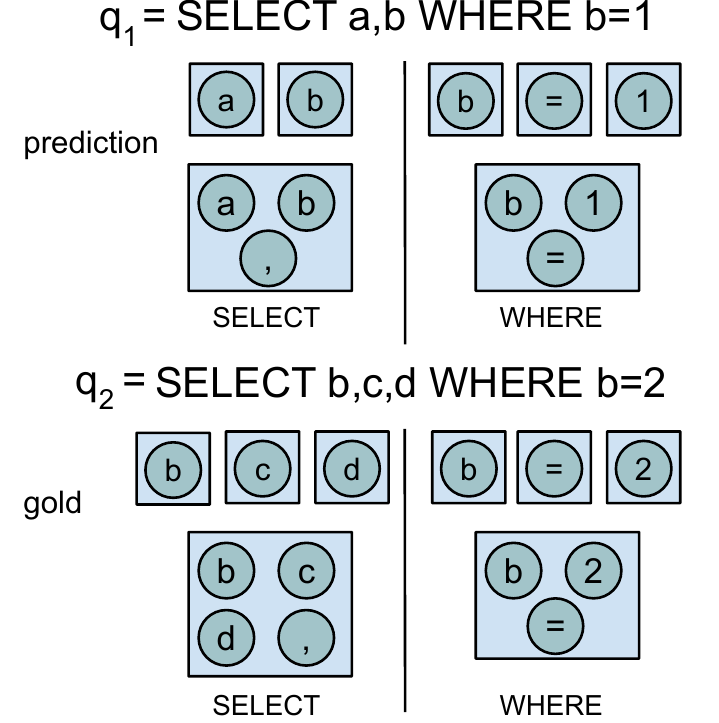}
  \caption{ An example for sub-tree matching.}
  \label{fig:metric}
\end{figure}

\begin{table*}[t]
  \small
  \centering
  \setlength{\tabcolsep}{2.5pt}
      \begin{tabular}{lccccc}
    \toprule
    \multicolumn{1}{c}{\bf Model} &
    \multicolumn{5}{c}{\bf Spider-\texttt{Dev}} \\
    \cmidrule(lr){2-6}
    &
    \multicolumn{1}{c}{PCM-F1} &
    \multicolumn{1}{c}{PCM-EM} &
    \multicolumn{1}{c}{PCM-F1-\textsc{NoValues}} &
    \multicolumn{1}{c}{PCM-EM-\textsc{NoValues}} &
    \multicolumn{1}{c}{EM}
    \\
    \midrule
    RAT \cite{wang2020ratsql} & 88.1 & 37.3 & 91.3 & 69.0 & 69.7\textsuperscript{$\dagger$} \\
    RAT+GAP \cite{shi2020learning} & 89.3 & 39.0 & 92.6 & 71.8 & 71.8\textsuperscript{$\dagger$} \\
    \midrule
    T5-Base \emph{with schema} & 85.7 & 56.7 & 85.9 & 57.2 & 57.6 \\
    T5-Large \emph{with schema} & \bf{86.3} & \textbf{61.2} & \textbf{86.6} & \textbf{62.6} & \bf{63.2} \\
    \bottomrule
  \end{tabular}
  \caption{Results on Spider with various metrics. While we do not focus on Spider, we show our results for comparison of the model and evaluation metric with a known benchmark. $\dagger$ denotes reported numbers from Spider's official leaderboard. PCM-F1-\texttt{NoValues} and PCM-EM-\texttt{NoValues} are modified versions of PCM-F1 and PCM-EM, respectively, such that all values in the SQL are anonymized and the \texttt{ON} clause is ignored, in order to compare with Spider's official Exact-Match (EM) metric.
  } \label{tab:results-spider}
\end{table*}

\subsection{Sub-tree elements matching}
In this work, in order to alleviate the aforementioned issues with exact-match logical form evaluation, we loosen it so that models can get partial scores if at least some part of their predicted expressions are found in the gold query. We do this by parsing both the predicted query and the gold query, comparing different parts of the two parsed trees and aggregating the scores into a single metric, as defined next. We term this metric \textbf{P}artial \textbf{C}omponent \textbf{M}atch \textbf{F1} (\textsc{PCM-F1}).

Our proposed metric is based on the ``Component Matching'' metric which is used in Spider's evaluation \cite{yu2019spider}, except that we use a parser that supports a large variety of queries (Spider's parser only supports specific types of queries), define how to compute the metric in a general way (not specific to any SQL-specific clause) and aggregate (average) the F1 scores into a single value, as defined next.

We first use an open-source SQL parser,  JSqlParser,\footnote{\url{https://github.com/JSQLParser/JSqlParser}} to parse a given SQL query $q$ into a tree, and extract a set of elements for each of its sub-trees, considering a sub-tree only if all of its leaves are terminal values in the query (similar to extracting constituents from a parse tree). For example, as can be seen in Figure~\ref{fig:metric}, the predicted query $q_1$ has 7 relevant sub-trees (marked in rectangles). The sub-tree which represents the expression \texttt{b=1} contains four elements: \texttt{b,=,1} and \texttt{b=1}. We then split these sets into different categories, based on the SQL query part that the root of the original sub-tree belonged to, for each of the following categories: $C=\{$\texttt{SELECT}, \texttt{TOP}, \texttt{FROM}, \texttt{WHERE}, \texttt{GROUPBY}, \texttt{HAVING}, \texttt{ORDERBY}$\}$. We denote all sets of elements for a query $q$ in a category $c \in C$ as $s_c(q)$. For example, as can be seen in Figure~\ref{fig:metric}, the clause $s_{\texttt{SELECT}}(q_1)$ yields 3 sub-trees. Given a predicted query $q_p$ and a gold query $q_g$, we compute the average F1 metric of all aligned pairs of sets $s_{c}(q_p)$ and $s_{c}(q_g)$:
$$ \textrm{PCM-F1}(q_p, q_g) = \frac{1}{\mid C \mid} \sum_{c\in C}{F1\left(s_c(q_p),s_c(q_g)\right)} $$
where F1 score is the harmonic mean of the precision and recall of the predicted sub-trees $s_c(q_p)$ with respect to the gold sub-trees $s_c(q_g)$. If for some category $c$, we get that $s_{c}(q_p)$ is an empty set but $s_{c}(q_g)$ is not, or vice-versa, we set $F1 = 0.0$ for that category.

Consider Figure~\ref{fig:metric} for an example. $s_{\texttt{SELECT}}(q_1)$ has 3 sub-trees while the gold category $s_{\texttt{SELECT}}(q_2)$ has 4 sub-trees. The predicted \texttt{SELECT} clause has 2 wrong sub-trees  (\texttt{a} and \texttt{a,b}) leading to a precision $p = \frac{1}{3}$, and 2 missing elements leading to a recall $r = \frac{1}{4}$. Similarly, the \texttt{WHERE} clause gets a precision of $p = \frac{1}{2}$ and a recall of $r = \frac{1}{2}$. Thus, we get $F1 = 0.285$ for $\texttt{SELECT}$ and $F1 = 0.5$ for \texttt{WHERE}, leading to a final score $ \textrm{PCM-F1} = 0.392$.

\subsection{Limitations}
\paragraph{Parsing Queries} JSqlParser could only parse 93.2\% of the validation SQL queries in \sede{}, and 92.5\% of the test queries. For that reason, for evaluation we only use the subset of queries which we can parse and evaluate \footnote{While we did not use the rest of the validation queries, we have released them in the dataset for future use, assuming at least some of them are valid queries.}. During evaluation, if the predicted query was not parsed, it receives a score of 0. Note that this does not affect training. 

\paragraph{False negatives}
We note that our metric does not address at all the issue of false negatives - in fact, since it's a looser metric than the Exact Match metric, it is actually more prone to produce false negative outcomes. For \sede{}, this issue could be mitigated by improving the similarity function that compares two queries, or by adapting the execution accuracy method in a way that will be less sensitive to instances of under-specification. We leave this challenge for future work.
\section{Experiments}

In this section, we describe our experimental setup, test how strong baselines perform on \sede{}, and analyze their errors.

\begin{table}
  \small
  \centering
  \setlength{\tabcolsep}{2.5pt}
      \begin{tabular}{lcccc}
    \toprule
    \multicolumn{1}{c}{\bf Model} &
    \multicolumn{2}{c}{\bf SEDE-\texttt{Dev}}  &
    \multicolumn{2}{c}{\bf SEDE-\texttt{Test}} \\
    \cmidrule(lr){2-3}
    \cmidrule(lr){4-5}
    &
    \multicolumn{1}{c}{PCM-F1} &
    \multicolumn{1}{c}{PCM-EM} &
    \multicolumn{1}{c}{PCM-F1} &
    \multicolumn{1}{c}{PCM-EM}
    \\
    \midrule
    T5-Base & 46.8 & \bf{4.0} & 49.4 & 3.6 \\
    ~~~\emph{with schema} & 46.4 & 3.4 & 48.9 & \bf{4.5} \\
    T5-Large & \bf{48.2} & \bf{4.0} & 50.6 & 4.1 \\
    ~~~\emph{with schema} & 47.1 & 3.7 & \bf{51.0} & 3.3 \\
    \bottomrule
  \end{tabular}
  \caption{Results on \sede{} development and test sets.
  } \label{tab:results}
\end{table}

\subsection{Experimantal Setup}
Most models in the Spider leaderboard\footnote{\url{https://yale-lily.github.io/spider}} use a grammar-based decoder designed for Spider, and as a result, they cannot be used as-is on \sede{}, which uses a larger grammar. Thus, following \citet{shaw2020compositional}, we use a general-purpose pre-trained sequence-to-sequence model, \textsc{T5} \cite{raffel2020exploring}, which was shown to be competitive with Spider's state-of-the-art models.

\begin{table*}[tb]
\resizebox{\textwidth}{!}{%
  \small
  \centering
  \setlength{\tabcolsep}{7pt}
      \begin{tabular}{m{3.2cm}m{6.5cm}m{6.5cm}}
    \toprule
    \multicolumn{1}{c}{\textbf{Category} \& Utterance} &
    \multicolumn{1}{c}{\bf Gold}  &
    \multicolumn{1}{c}{\bf Predicted}
    \\
    \midrule
    \textbf{Under-specification and Hidden assumptions} \newline\newline
    \emph{Positive scored questions without answers for c++ tags} & \makecell[llllllll]{
    \texttt{SELECT id as "" Post Link "", *} \\ \texttt{FROM posts} \\
    \texttt{WHERE} \\
    \texttt{~~~answercount = 0} \\
    \texttt{~~~AND tags NOT LIKE '\%c++\%'} \\
    \texttt{~~~AND score > 0} \\
    \texttt{~~~AND ClosedDate is null} \\
    \texttt{ORDER BY score DESC}} & \makecell[llllllllll]{
    \texttt{SELECT id as [Post Link], tags,} \\
    \texttt{~~~~score, viewcount, CreationDate} \\
    \texttt{FROM posts} \\
    \texttt{WHERE} \\
    \texttt{~~~tags NOT LIKE '\%c++\%'} \\
    \texttt{~~~AND answercount = 0} \\
    \texttt{~~~AND posttypeid = 1} \\
    \texttt{~~~AND score > 0} \\
    \texttt{~~~AND CreationDate > '2018-01-01'} \\
    \texttt{ORDER BY CreationDate DESC}} \\
    \midrule
    \textbf{Dates Manipulation} \newline\newline
    \emph{Percentage of votes - depending on day after posting (only questions)} & \makecell[llllllll]{
    \texttt{SELECT} \\ \texttt{~~~DATEDIFF(day, p.CreationDate,} \\
    \texttt{~~~~~~v.CreationDate) AS Days,} \\
    \texttt{~~~COUNT(v.Id) AS Count,} \\
    \texttt{~~~COUNT(v.Id) * 100.0} \\
    \texttt{~~~~~~/ SUM(COUNT(v.Id)) OVER ()} \\
    \texttt{~~~~~~AS Percentage ...}} & \makecell[llllll]{
    \texttt{SELECT} \\ \texttt{~~~DATEDIFF(day, p.CreationDate,} \\
    \texttt{~~~~~~v.CreationDate) AS Days,} \\
    \texttt{~~~COUNT(v.Id) AS Count,} \\
    \texttt{~~~COUNT(v.Id) * 100.0} \\
    \texttt{~~~~~~/ COUNT(v.Id)} \\
    \texttt{~~~~~~AS Percentage ...}} \\
    \midrule
    \textbf{Parameters} \newline \newline
    \emph{Top users in a tag by score and answer count} & \makecell[lll]{
    \texttt{SELECT TOP 100 ...} \\ \texttt{WHERE ...} \\ \texttt{~~~AND t.TagName = '\#\#tagName\#\#'}} & \makecell[lll]{
    \texttt{SELECT TOP \#\#num?100\#\# Users.id} \\ \texttt{~~~AS [User Link], ...} \\
    \texttt{WHERE tags.TagName = '\#\#tagname\#\#'}} \\
    \bottomrule
  \end{tabular}}
  \caption{Error analysis of gold queries vs. predicted queries for some selected dataset characteristics mentioned in \ref{sec:dataset}. For brevity, in some of the examples we show only relevant parts of the query.} \label{tab:analysis}
\end{table*}

Since all queries in \sede{} come from a single schema which is seen during training time, it is not clear if allowing the model to access the schema during encoding and decoding is helpful. We thus experiment with two versions. In the first one, \textsc{T5}, the input is simply the utterance $\bar{u}$. In the second, \textsc{T5} \emph{with schema}, the input is the utterance $\bar{u}$ followed by a separator token, and then the serialized schema. We follow \citet{suhr-etal-2020-exploring} and serialize the schema by listing all tables in the schema and all the columns for each table, with a separator token between each column and table. Naturally, we did not evaluate \textsc{T5} 
(without schema) on Spider since encoding the schema is crucial in a zero-shot setup.
We perform textual pre-processing to the queries in \sede{} before training (i.e. remove non UTF-8 characters and SQL comments, normalize spaces and new lines, normalize apostrophes, remove comments, etc.).
We show results for experiments considering the titles alone, and ignore their given description, which are given in 14.6\% of the examples. We have found that if we concatenate the description to the title, we get slightly worse results.

We use the SentencePiece \cite{kudo-richardson-2018-sentencepiece} tokenizer, with its default vocabulary, for all models. We fine-tune the model to minimize the token-level cross-entropy loss against the gold SQL query for 60 epochs with the AdamW \cite{loshchilov2019decoupled} optimizer and a learning rate of $5e^{-5}$. We choose the best model based on the performance on the validation set for each dataset, using Exact-Match (\textsc{EM}) for Spider and \textsc{PCM-F1} for \sede{}. For inference, we use beam-search (of size 6) and choose the highest-probability generated SQL query. We show results for both \textsc{T5}-Base and \textsc{T5}-Large.

For each experiment we measure \textsc{PCM-F1} together with a modified version of it, \textsc{PCM-EM} (PCM exact match), that returns an accuracy of 1 for a given prediction if and only if the \textsc{PCM-F1} value for that prediction is 1. For Spider, we use the officially provided script to measure the \textsc{EM} metric.

\subsection{Main Results}
\label{subsec:results}
We show experiments results for \sede{} in Table~\ref{tab:results} and for Spider in Table~\ref{tab:results-spider}. The results indicate that the performance gap between \sede{} and Spider is large: while T5-Large reaches a score of 63.2 \textsc{EM} on Spider's validation set, not very far from the state-of-the-art (a difference of 8.6 points), and a \textsc{PCM-F1} of 86.3, when trained on \sede{}, it only receives 48.2 and 50.6 \textsc{PCM-F1} on the validation and test set of \sede{}, respectively. This supports our main claim, that single-schema datasets could still impose a substantial challenge when tested in a realistic setup.

We also notice in Table~\ref{tab:results-spider} that large improvements in \textsc{EM} do not necessarily imply a large increase in \textsc{PCM-F1}. Since the \textsc{PCM-F1} numbers are high ($>$85\%) for Spider in any of the tested models, we conclude that in Spider, in most cases the models are generating SQL queries that are similar in their high-level structure to the gold SQL, but are only different by smaller changes (e.g. value or column name).

Comparing experiments with and without encoding the schema shows that encoding the schema does not significantly improve results in this single-domain setup.
We also observe that \textsc{PCM-EM} is close to 0 in all experiments, supporting our motivation to create a loosened evaluation metric.

\subsection{PCM-F1 Validation}
\label{subsec:correctnes}
In order to validate the correctness of our proposed evaluation metric, we compare \textsc{PCM-EM} with the more established \textsc{EM} metric of Spider. 
There are two differences in the way \textsc{EM} is calculated compared to \textsc{PCM-EM}: (1) \textsc{EM} anonymizes all values in the queries and (2) \textsc{EM} ignores the \texttt{ON} expressions in the \texttt{JOIN} clauses. For these reasons, we define PCM-F1-\textsc{NoValues} and \textsc{PCM-EM}-\textsc{NoValues}, modified versions of \textsc{PCM-F1} and \textsc{PCM-EM}, respectively, such that all values in the SQL are anonymized and the \texttt{ON} expressions are ignored. Table~\ref{tab:results-spider} shows that \textsc{EM} and \textsc{PCM-EM}-\textsc{NoValues} are only different by up to 0.7 points for all models, showing that \textsc{PCM-F1} and \textsc{PCM-EM} are well calibrated with Spider's \textsc{EM}.

\subsection{Error Analysis}

Next, we analyze errors and successful outputs of the model. Table~\ref{tab:analysis} shows examples of gold vs. predicted queries by our model, with respect to some of the introduced challenges mentioned in \ref{subsec:character}.

We can see from the first example that the model is often wrong whenever the question is not specified well: In this example, this happens in the \texttt{SELECT}, \texttt{WHERE} and \texttt{ORDER} fields. In the \texttt{SELECT} clause, the model predicts extra columns in comparison to the gold query, most likely as it has learned to do so for similar questions. In addition, since the desired order of the results are not mentioned in the utterance, it leads to a different predicted \texttt{ORDER BY} clause. A hidden assumption the author had added to the query is taking into account only open questions (i.e. questions with no close date: \texttt{ClosedDate is null}). The model, which could not deduce this assumption from the utterance alone, predicts a wrong filter expression \texttt{CreationDate > '2018-01-01'}.

The second example shows how the model correctly uses the \texttt{DATEDIFF} function to manipulate dates, although it predicted a wrong computation of the percentage (i.e. without the \texttt{SUM} function).

The last example shows how the model generates a SQL query with parameters, for the number of required users (with a predicted default value of 100) and for the tag name. In this case, the predicted query is possibly better than the gold one as it uses a reusable parameter instead of a fixed one.
\section{Conclusion}

In this work, we take a significant step towards improving and evaluating Text-to-SQL models in a real world setting, by releasing \sede{}, a dataset comprised of real-world complex and diverse SQL queries with their utterances, naturally written by real users. We show that there's a large gap between the performance of strong Text-to-SQL baselines on \sede{} compared to the commonly studied dataset Spider, and hope that the release of this challenging dataset will encourage research on improving generalization for real-world SQL prediction.

\paragraph{Acknowledgments}

We thank Kevin Montrose and the rest of the Stack Exchange team for providing the raw query log.

\bibliographystyle{acl_natbib}
\bibliography{anthology,acl2021}


\end{document}